\documentclass[10pt,letterpaper]{article}
\usepackage{xcolor}
\usepackage[top=0.85in,left=2.75in,footskip=0.75in,marginparwidth=2in]{geometry}

\usepackage[utf8]{inputenc}

\usepackage{cite}

\usepackage{nameref}
\usepackage{hyperref}
\hypersetup{
    colorlinks=true,    
    linkcolor=blue,     
    citecolor=blue,     
    urlcolor=blue,      
    filecolor=blue,     
    pdfborder={0 0 0}   
}

\usepackage{microtype}
\DisableLigatures[f]{encoding = *, family = * }

\usepackage{changepage}

\usepackage[aboveskip=1pt,labelfont=bf,labelsep=period,singlelinecheck=off]{caption}

\makeatletter
\renewcommand{\@biblabel}[1]{\quad#1.}
\makeatother

\usepackage{lastpage,fancyhdr,graphicx}
\usepackage{epstopdf}
\fancyheadoffset[L]{2.25in}
\fancyfootoffset[L]{2.25in}

\usepackage{color}

\definecolor{Gray}{gray}{.25}

\usepackage{graphicx}

\usepackage{sidecap}

\usepackage{wrapfig}
\usepackage[pscoord]{eso-pic}
\usepackage[fulladjust]{marginnote}
\reversemarginpar

\usepackage{amsmath,amssymb,amsfonts}
\usepackage{amsthm}
\usepackage{mathrsfs}
\usepackage{booktabs}
\usepackage{multirow}
\usepackage{algorithm}
\usepackage{algorithmicx}
\usepackage{algpseudocode}
\usepackage{listings}

\theoremstyle{plain}

\begin{document}
\vspace*{0.35in}

\begin{flushleft}
{\Large
\textbf\newline{A Method for the Architecture of a Medical Vertical Large Language Model Based on Deepseek R1}
}
\newline
\\
Mingda Zhang\textsuperscript{1},
Jianglong Qin\textsuperscript{1,2*}
\\
\bigskip
\bf{1} School of Software, Yunnan University, Kunming, 650500, China
\\
\bf{2} Yunnan Key Laboratory of Software Engineering, Yunnan University, Kunming, 650500, China
\\
\bigskip
* qinjianglong@ynu.edu.cn

\end{flushleft}

\section*{Abstract}
Despite significant advances in foundation models like DeepSeek-R1 and ChatGPT, their deployment in medical settings faces critical challenges including computational requirements and professional knowledge barriers. This paper presents an efficient lightweight medical large language model architecture that systematically addresses these challenges through three-dimensional optimization: knowledge acquisition, model compression, and computational enhancement. We design a knowledge transfer pipeline from DeepSeek-R1-Distill-70B to DeepSeek-R1-Distill-7B using Low-Rank Adaptation (LoRA) for precise medical knowledge retention. Through 4-bit quantization and mixed-precision strategies, we achieve substantial model compression while preserving medical reasoning capabilities. The inference framework incorporates Flash Attention acceleration and continuous batching, complemented by specialized prompt templates for diverse medical queries. Experimental evaluation on medical benchmarks demonstrates that our approach maintains 92.1\% accuracy on USMLE examinations while reducing memory consumption by 64.7\% and inference latency by 12.4\% compared to baseline models. This work provides a practical solution for deploying advanced language models in resource-constrained medical environments, enabling broader accessibility of AI-assisted healthcare.


\section*{Introduction}

Medical large language models \cite{bib1,bib2,bib3,bib4,bib5} represent an important research direction at the intersection of artificial intelligence and healthcare. Currently, most research focuses on the direct application or simple fine-tuning of general large models in medical scenarios. However, in actual medical environments, model deployment faces multiple challenges including professional knowledge barriers, computational resource limitations, and deployment environment constraints. General large models requiring hundreds of billions of parameters and high-performance computing equipment are difficult to meet the practical needs of primary healthcare institutions. This paper investigates lightweight medical vertical large language model construction methods based on knowledge distillation \cite{bib6}, aiming to significantly reduce computational resource requirements while maintaining professional accuracy, with broad application prospects in clinical auxiliary diagnosis \cite{bib7}, medical Q\&A \cite{bib8}, medical education \cite{bib9,bib10}, and other fields.

In recent years, knowledge distillation and model compression technologies (such as LoRA \cite{bib11}) have improved the deployment efficiency of large language models, enabling them to run in resource-constrained environments. Due to their excellent parameter efficiency and performance retention capabilities, more and more researchers are applying these technologies to the medical artificial intelligence field. Currently, most works \cite{bib12,bib13,bib14} reduce model computational complexity and improve inference speed through parameter-efficient fine-tuning or model quantization. For vertical scenarios with high medical professional requirements, Chen et al. \cite{bib6} attempted to apply knowledge distillation technology to medical large language model construction, transferring medical knowledge from large-scale teacher models to small-scale student models. Although these methods solve the model deployment efficiency problem to some extent, they often ignore the synergistic effects of knowledge acquisition, model compression, and computational optimization across three dimensions, lacking a systematic lightweight solution. In complex scenarios such as medical Q\&A, multi-dimensional collaboration plays a more important role in building efficient medical large language models.

This paper proposes an efficient lightweight medical vertical large language model construction method based on a three-dimensional collaborative strategy. First, a knowledge transfer pipeline is designed from a large-scale teacher model (70B) to a lightweight student model (7B), and Low-Rank Adaptation (LoRA) fine-tuning is adopted to acquire professional medical knowledge. Then, model compression techniques including 4-bit weight quantization are implemented. Finally, inference optimization strategies such as Flash Attention acceleration are integrated to improve operational efficiency. Specifically, in the knowledge acquisition dimension, a medical Q\&A dataset is constructed through design questions such as "Please analyze in detail the possible diagnosis and treatment plans for a patient with \{symptom description\}", combining teacher model outputs with professional medical knowledge bases to build training corpus; in the model compression dimension, low-rank decomposition and quantization techniques are applied to significantly reduce model parameter size; in the computational optimization dimension, attention mechanism acceleration algorithms and batching strategies adapted to medical scenarios are developed, ultimately building a lightweight medical large language model with high computational efficiency and strong professionalism.

The main contributions of this paper include 3 aspects:

\begin{enumerate}
\item Proposes a three-dimensional collaborative lightweight medical vertical large language model construction method, systematically integrating knowledge acquisition, model compression, and computational optimization technologies, improving model deployment efficiency while ensuring professional accuracy.

\item Designs a knowledge distillation and parameter-efficient fine-tuning framework specifically targeting medical scenarios, achieving efficient retention of medical professional capabilities in lightweight models through combining teacher model knowledge and LoRA fine-tuning, enhancing model performance in complex medical Q\&A scenarios.

\item Conducts extensive experiments on multiple standard medical Q\&A datasets. Experimental results show that compared to existing methods, the lightweight medical large language model proposed in this paper maintains professional capabilities while reducing memory consumption by 64.7\% and inference latency by 12.4\% compared to directly fine-tuned DeepSeek-R1-7B (FP16) models; while ensuring 92.1\% accuracy in the United States Medical Licensing Examination (USMLE) test.
\end{enumerate}

\section*{Related Work}

\subsection*{Knowledge Acquisition Technologies for the Medical Field}

Medical domain language model research has evolved along two technical paths: specially trained medical models and domain adaptation of general large models \cite{bib15}. Dedicated models such as Med-PaLM 2 \cite{bib16} and GatorTron \cite{bib17}, although they have built rich domain knowledge representations through pretraining on medical texts, generally face systemic challenges including barriers to medical data acquisition, excessive computational resource requirements, and difficulties in coordinating professional and general capabilities \cite{bib15,bib18}. These inherent limitations have prompted researchers to explore more efficient medical knowledge transfer methods.

The hierarchical knowledge acquisition architecture proposed in this paper innovatively integrates adaptive LoRA strategies with a two-stage knowledge transfer framework, achieving efficient medical knowledge transfer from 70B parameters to 7B parameters. The key technological innovations of this architecture are reflected in three aspects: First, expanding the LoRA target module range to key components of the feed-forward network in the DeepSeek architecture, allowing the model to more comprehensively adapt to the professional language characteristics of the medical field; second, introducing Rank-Stabilized Low-Rank Adaptation (RSLoRA) technology, reducing the error rate of medical terminology representation by optimizing the gradient update mechanism; finally, designing a progressive batch distillation strategy, constructing training sequences according to the gradient of medical concept complexity, enhancing the completeness and accuracy of knowledge transfer \cite{bib19}. This knowledge acquisition strategy not only promotes efficient transfer of clinically relevant information but also reduces computational resource requirements, laying the foundation for subsequent model compression.

\subsection*{Model Compression Methods for Medical Language Models}

Model compression technology constitutes the core link in the lightweight implementation of medical large language models. Existing medical domain compression methods such as DistillBERT-Med \cite{bib20} and MiniMedLM \cite{bib21} mainly rely on network structure optimization to reduce parameter scale, but often exhibit significant professional degradation in complex medical scenarios. Particularly in parameter quantization technology applications, although INT8/INT4 quantization can significantly reduce computational resource requirements, direct application to medical models often leads to serious loss of clinically critical information such as drug dosage accuracy decreases and professional terminology recognition errors \cite{bib22}, a problem that is particularly prominent in resource-constrained environments.

To address these challenges, the mixed-precision quantization and hierarchical device mapping technology designed in this paper achieves differentiated resource optimization. This method achieves synergistic improvement of resource optimization and medical precision through three technical innovations: First, applying a differentiated quantization strategy, maintaining key attention mechanisms at 8-bit precision while quantizing feature extraction layers to 4-bit precision using the NF4 method optimized for medical terminology \cite{bib23}; second, constructing a device affinity optimization process, systematically analyzing the computational complexity of each layer and establishing a device-layer affinity matrix to achieve optimal allocation of computational resources; third, implementing a dynamic load balancing mechanism, adjusting resource allocation strategies in real-time according to inference requirements \cite{bib24}. This hierarchical compression strategy tailored to medical model characteristics significantly reduces resource requirements while maintaining professional knowledge expression capabilities, creating conditions for further optimization of computational efficiency.

\subsection*{Computational Optimization Strategies for Large Medical Models}

Computational optimization strategies solve the efficiency and performance balance dilemma in the deployment of large medical models. Existing technologies such as Flash Attention \cite{bib25} and CUDA graph optimization \cite{bib26} have made certain progress in general domains, but lack targeted optimization for characteristics specific to medical models such as long context understanding, intensive reasoning with professional terminology, and differentiated medical problem processing. Furthermore, despite breakthroughs in multi-engine integration technology, the lack of adaptive deployment mechanisms for different clinical scenarios such as emergency triage and complex case analysis limits the application flexibility and performance of models in actual medical environments.

This paper proposes the Model Inference Engine Optimization (MIEO) framework, establishing a unified theoretical framework for inference optimization of medical language models. Based on MIEO theory, this paper establishes a hierarchical three-layer optimization architecture: the computation acceleration layer implements shape-aware caching and low-precision acceleration by identifying the clustering characteristics of medical problem input shapes; the resource scheduling layer integrates a two-level cache system and asynchronous streaming architecture; the engine abstraction layer provides multi-engine integration and environment awareness capabilities, adaptively selecting optimal inference strategies based on hardware configuration and runtime conditions; the domain adaptation layer designs dedicated optimization components for medical problem classification and terminology standardization \cite{bib27}. This collaborative optimization of the multi-layer architecture achieves dual enhancement of medical professionalism and computational efficiency, providing a systematic solution for medical AI deployment in resource-constrained environments while ensuring knowledge acquisition quality and professional performance after model compression.

\section*{Medical Large Language Model Core Technical Architecture}

\subsection*{Task Objectives and Overall System Architecture}

This paper aims to build a high-performance, resource-efficient medical large language model system capable of providing professional, accurate medical Q\&A services in resource-constrained environments. Core objectives include: (1) effectively acquiring and applying domain-specific medical knowledge; (2) achieving efficient model deployment in computationally resource-constrained environments; (3) optimizing inference performance to ensure response timeliness and professionalism.

The medical large language model system proposed in this paper consists of three main technical modules: hierarchical knowledge acquisition architecture, resource-efficient transformation and deployment technology, and high-performance inference optimization framework. These three modules form a complete technical link, from medical knowledge acquisition and storage to efficient inference generation, together constituting the technical contribution of this paper.

\begin{figure}[ht]
\centering
\includegraphics[width=0.85\textwidth]{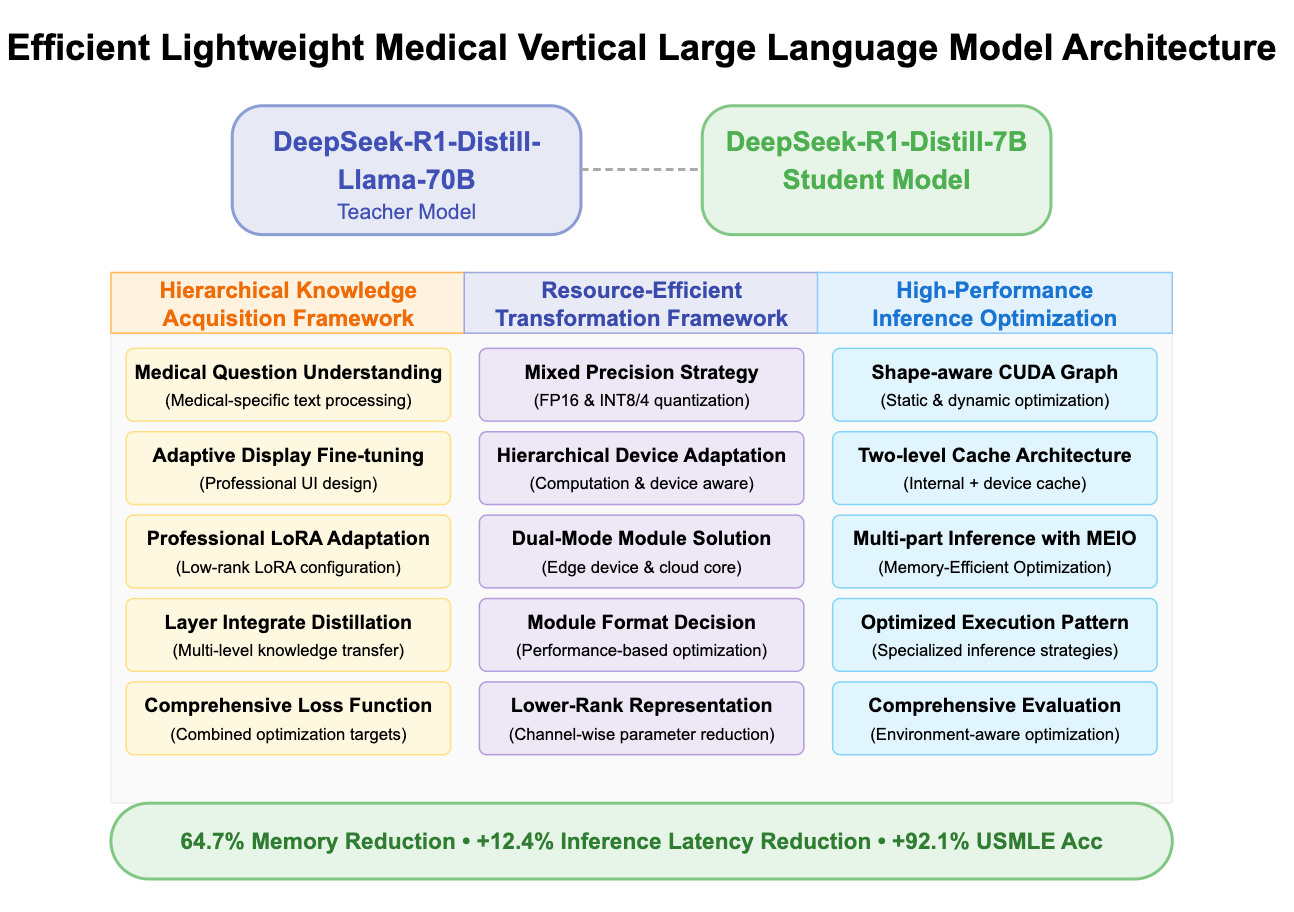}
\caption{\textbf{Efficient Lightweight Medical Vertical Large Language Model Architecture Method.} This figure illustrates the three-dimensional collaborative architecture integrating hierarchical knowledge acquisition, resource-efficient transformation, and high-performance inference optimization.}
\label{fig1}
\end{figure}

The experimental steps first involve medical knowledge acquisition, completing efficient knowledge acquisition and transfer through medical problem classification, professional prompt template design, adaptive LoRA fine-tuning, and large-scale knowledge distillation; then resource-efficient transformation, applying mixed-precision quantization and hierarchical device mapping technology to reduce model resource requirements and implement dual-mode deployment solutions; finally designing a high-performance inference optimization framework, achieving efficient model operation in various hardware environments through the multi-dimensional inference performance optimization model. The following sections will detail the design principles and implementation methods of these technical modules.

\subsection*{Hierarchical Knowledge Acquisition Architecture}

The hierarchical knowledge acquisition architecture proposed in this paper achieves effective acquisition and application of medical knowledge through a systematic method, integrating cross-disciplinary knowledge from natural language processing and the medical professional domain to form a complete knowledge acquisition framework.

The medical problem classification system proposed in this paper adopts a keyword-based hierarchical classification method, dividing medical problems into five main categories (medication information, diagnosis, treatment plans, preventive healthcare, and emergency treatment). For a problem $q$, the function $C(q)$ maps it to the corresponding category $c_i$:

\begin{equation}
C(q) = \arg{\max_{c_{i} \in \mathcal{C}}{\sum_{w \in q}^{}{I\left( w \in K_{c_{i}} \right)}}}
\end{equation}

where $c_i$ represents the set of all categories, $K_{c_i}$ is the set of keywords for category $c_i$, $I$ is an indicator function that takes the value 1 when word $w$ belongs to the keyword set of category $c_i$, and 0 otherwise. This classification function selects the category with the highest matching degree as the classification result for the problem $q$ by calculating the matching degree of the problem $q$ with the keywords of each category.

Based on the classification results, a dedicated prompt template $P_{c_i}$ is designed for each type of problem:

\begin{equation}
P_{c_{i}}(q) = P_{\text{base}} \oplus S_{c_{i}}
\end{equation}

where $P_{\text{base}}$ is a basic prompt containing medical professional guidance principles, $S_{c_i}$ is a specific suffix for category $c_i$, and $\oplus$ represents text concatenation operation. In the experiments, we first constructed a basic prompt including medical professional standards and evidence-based medical principles, then added specific guiding suffixes according to the problem category, such as emphasizing timeliness and step clarity for emergency problems, and dose accuracy and contraindication explanations for medication problems \cite{bib29}. This classification-driven prompt design method improves the efficiency of professional medical knowledge acquisition.

This paper proposes an Adaptive Low-Rank Adaptation (ALORA) fine-tuning strategy, extending the standard LoRA method to significantly reduce additional parameter quantity and improve computational efficiency while maintaining model performance. Its core idea is to update weights only in a low-rank subspace: given an input vector $x$, the output vector $h$ can be expressed as:

\begin{equation}
h = W_{0}x + \Delta Wx = W_{0}x + \alpha rWx
\end{equation}

where $W_0 \in R^{d \times k}$ is the original weight matrix of the model, $d$ and $k$ represent the row and column dimensions respectively; $\Delta W = BA$ is the weight update in low-rank decomposition form (representing an "increment" on the original weights), $B \in R^{d \times r}$ and $A \in R^{r \times k}$ are both learnable small matrices; $r$ is a low-rank parameter used to control the parameter scale during fine-tuning; $\alpha$ is a scaling factor used to balance the influence of fine-tuning on the original weights. By "superimposing" this low-rank increment on the original weights $W_0$, ALORA captures new task features without significantly increasing parameter overhead, providing a feasible solution for efficiently fine-tuning large models \cite{bib11}.

This paper further introduces Rank-Stabilized LoRA (RSLoRA) technology, improving numerical stability by modifying the gradient update mechanism:

\begin{equation}
\Delta W = \text{StableSVD}(BA) = \sum_{i = 1}^{r}{\sigma_{i}u_{i}v_{i}^{T}} + \lambda I
\end{equation}

where $\text{StableSVD}$ is a stabilized singular value decomposition operation, $\sigma_i$ is a singular value, $u_i$ and $v_i$ are the corresponding singular vectors, $\lambda$ is an adaptive regularization term, and $I$ is the identity matrix. In the experiments, $\lambda$ is dynamically adjusted according to the matrix condition number, truncated SVD technology is used to filter numerically insignificant components, and a gradient scaling mechanism is implemented to prevent gradient explosion \cite{bib30}. This stabilization mechanism is particularly suitable for the precise acquisition of professional knowledge such as medical terminology.

This paper designs a two-stage knowledge acquisition framework, including supervised fine-tuning and knowledge distillation. The fine-tuning process adopts an adaptive learning rate scheduling strategy:

\begin{equation}
\eta_{t} = \eta_{\min} + \frac{1}{2}\left( \eta_{\max} - \eta_{\min} \right)\left( 1 + \cos\left( \frac{t \cdot \pi}{T \cdot R} \right) \right)
\end{equation}

where $\eta_t$ is the learning rate at step $t$, $\eta_{\min}$ is the minimum learning rate, $\eta_{\max}$ is the maximum learning rate, $T$ is the total number of training steps, and $R$ is the number of restarts. This cosine annealing with restart strategy enables the model to escape local optima and explore a wider parameter space \cite{bib31} by periodically reducing and restoring the learning rate, improving the completeness of medical knowledge acquisition.

The knowledge distillation stage adopts a comprehensive loss function:

\begin{equation}
\mathcal{L} = \lambda_{1}\mathcal{L}_{\text{CE}} + \lambda_{2}\mathcal{L}_{\text{KL}} + \lambda_{3}\mathcal{L}_{\text{MSE}} + \lambda_{4}\sum_{e \in E}w_{e}\left| P_{S}(e) - P_{T}(e) \right|
\end{equation}

where $\mathcal{L}_{\text{CE}}$ is the cross-entropy loss, ensuring consistency between student model predictions and true labels; $\mathcal{L}_{\text{KL}}$ is the divergence loss, measuring the overall distribution difference between student model outputs and teacher model outputs; $\mathcal{L}_{\text{MSE}}$ is the mean squared error loss, used to align the intermediate representations of the student model and teacher model; $E$ is the set of key medical entities, $w_e$ is the importance weight of entity $e$, $|P_S(e) - P_T(e)|$ represents the absolute difference between the prediction probabilities of the student model and teacher model for a key medical entity $e$, i.e., measuring the degree of deviation between the two in entity prediction; $\lambda_1$, $\lambda_2$, $\lambda_3$, $\lambda_4$ are parameters controlling the weights of each loss term, automatically tuned through validation set performance. In the experiments, the optimal weight combination was determined through grid search, and the weights for key medical entities (such as drug dosages, contraindications, diagnostic criteria) were set to higher values, ensuring accurate transfer of medical professional knowledge \cite{bib32}. This multi-objective distillation strategy is crucial for medical knowledge acquisition, providing the foundation for maintaining high-quality medical reasoning capabilities in subsequent model compression.

\subsection*{Resource-Efficient Transformation and Deployment Technology}

The resource-efficient transformation and deployment technology described in this section, through mixed-precision quantization and hierarchical device mapping technology, can run efficiently on consumer-grade hardware and is the core technical link of model compression.

The mixed-precision quantization strategy maintains key attention mechanisms in the model at 8-bit precision, while feature extraction layers are quantized to 4-bit precision using the NF4 method \cite{bib33,bib38}. In the experiments, we first identified the components most sensitive to medical terminology and numerical expressions through attention weight analysis, then applied differentiated quantization strategies to different components. NF4 implementation optimizes quantization points through statistical analysis of weight distribution, ensuring accurate expression of clinically critical information \cite{bib33}. This differentiated quantization method is key to maintaining medical model professionalism after compression.

Hierarchical device mapping technology selectively allocates different network layers to heterogeneous computing resources based on their computational characteristics and memory requirements, achieving optimal allocation of computational resources. Experimental steps include: (1) analyzing the computational complexity and memory requirements of each layer; (2) constructing a device-layer affinity matrix; (3) applying integer linear programming to maximize inference efficiency. For example, computationally intensive input embedding layers and output projection layers are allocated to GPU processing, while intermediate transformer blocks are appropriately offloaded to CPU to optimize memory utilization \cite{bib34}. This hierarchical device mapping is the core technology for achieving efficient deployment after model compression.

The dual-mode deployment solution proposed in this paper includes adapter mode and merged mode, achieving adaptive scheduling of the system through a dynamic switching mechanism, organically integrating the advantages of both modes. The deployment mode decision function is implemented through the following formula:

\begin{equation}
D(M,E,R) = w_{1} \cdot P(M,E) + w_{2} \cdot F(M,R) - w_{3} \cdot C(M,E)
\end{equation}

where $D$ is a deployment performance evaluation function, $M$ represents the deployment mode (adapter mode or merged mode), $E$ represents the current system resource conditions, $R$ represents clinical needs; $P(M,E)$ represents the performance indicators (such as response time, throughput) of mode $M$ under resource conditions $E$; $F(M,R)$ measures the functional adaptation degree of mode $M$ to clinical needs $R$ (such as professional domain coverage, knowledge breadth); $C(M,E)$ quantifies the computational resource consumption of mode $M$; $w_1$, $w_2$, $w_3$ are weight coefficients, dynamically adjusted according to the medical institution's priorities. In the experiments, the optimal weight combination was determined through Bayesian optimization methods, and an adaptive weight adjustment mechanism based on historical decision performance was designed, enabling the system to continuously optimize deployment strategies according to actual application scenario needs \cite{bib35}. This adaptive deployment technology is an important means to solve flexible resource scheduling after model compression.

\subsection*{High-Performance Inference Optimization Framework}

This paper constructs a three-layer structure MIEO architecture: computation acceleration layer, resource scheduling layer, and engine abstraction layer. The experiments first conducted performance benchmark testing on various hardware platforms, then systematically evaluated different levels of optimization strategies, and finally built a complete inference optimization framework. Key technological innovations include:

\begin{enumerate}
\item Shape-aware CUDA graph caching mechanism: Eliminates repeated graph construction and compilation overhead by capturing and reusing optimized computation graphs. The implementation process includes: (1) capturing computation graphs of common input shapes during the warm-up phase; (2) continuously monitoring new shape patterns and dynamically expanding the cache during the runtime phase; (3) mapping approximate shapes to cached graphs through slight padding adjustments, improving cache hit rates \cite{bib36}. This caching mechanism is crucial for optimizing inference computation of medical models.

\item Memory-disk two-level cache and asynchronous streaming architecture: Significantly reduces average response time through hierarchical storage and retrieval mechanisms. The memory cache maintains response results for high-frequency queries, providing sub-millisecond access speed; the disk cache saves a wider range of historical responses. The system adopts a cache matching algorithm based on semantic similarity:

\begin{equation}
\text{Similarity}\left( q_{1},q_{2} \right) = \alpha \cdot \text{Jaccard}\left( q_{1},q_{2} \right) + (1 - \alpha) \cdot \text{Cosine}\left( \text{Embed}\left( q_{1} \right),\text{Embed}\left( q_{2} \right) \right)
\end{equation}

where $\text{Jaccard}(q_1,q_2)$ calculates the vocabulary overlap rate between two queries, $\text{Embed}(q)$ represents the semantic embedding representation of query $q$, $\text{Cosine}$ calculates cosine similarity, and $\alpha$ is a coefficient balancing vocabulary matching and semantic matching \cite{bib37}. This two-level cache structure is particularly suitable for medical reasoning optimization, significantly reducing computational latency.

\item Multi-engine integration and environment adaptive mechanism: Constructs an environment-aware multi-engine integration framework capable of dynamically selecting optimal inference strategies based on the running environment. The system builds a multi-dimensional decision model by real-time monitoring of hardware resources, software environment, and workload characteristics:

\begin{equation}
E^{*} = \arg{\max_{E_{i} \in E}S}\left( E_{i},H,L \right)
\end{equation}

where $E^*$ is the selected optimal engine configuration, $E$ is the set of all available engine configurations, $S$ is a comprehensive scoring function, $H$ represents hardware environment parameters, and $L$ represents workload characteristics. The system continuously updates environmental states during runtime and estimates the expected performance of different engine configurations through prediction models, thereby selecting an inference engine suitable for the current scenario \cite{bib35}. This adaptive mechanism significantly improves the resource utilization of the system on heterogeneous computing platforms, and minimizes latency and energy consumption while ensuring inference quality, suitable for resource-constrained or dynamically changing deployment environments.
\end{enumerate}

In summary, the high-performance inference optimization framework in this paper fully integrates professional capabilities after knowledge acquisition and lightweight characteristics after model compression, achieving efficient operation of medical large language models in resource-constrained environments through multi-level computational optimization technologies, providing key support for the final performance of the three-dimensional collaborative strategy.

\section*{Experimental Evaluation and Analysis}

\subsection*{Experimental Overview}

This paper designed a series of experiments to systematically verify the effectiveness of the proposed three-dimensional collaborative lightweight medical vertical large language model construction method. The experiments revolve around the three core dimensions of knowledge acquisition, model compression, and computational optimization, using multiple standard medical evaluation datasets, multiple baseline models, and different hardware environments to comprehensively evaluate the model's medical professional capability and resource efficiency. Through experimental designs in three aspects—professional accuracy comparison, resource efficiency analysis, and component contribution ablation—the paper verified the ability of the proposed method to reduce resource requirements while maintaining medical professionalism.

\subsection*{Experimental Results and Analysis}

\subsubsection*{Medical Professional Capability Assessment}

Table 1 shows the accuracy comparison of different models on the USMLE medical licensing examination. USMLE Step 1 accuracy is calculated as the proportion of correct answers in standardized medical tests, while professional knowledge coverage is measured by the matching degree with core concepts in the medical knowledge base.

\begin{table}[!ht]
\begin{adjustwidth}{-0.5in}{0in}
\caption{\textbf{Accuracy Comparison of Large Language Models on USMLE Medical Licensing Examination}}
\label{tab1}
\begin{tabular}{lcc}
\toprule
\textbf{Model} & \textbf{USMLE Step 1} & \textbf{Medical Professional} \\
 & \textbf{Accuracy (\%)} & \textbf{Knowledge Coverage (\%)} \\
\midrule
Claude-3.7 Sonnet Extended & 92.8 & 94.1 \\
DeepSeek-R1-Distill-7B-Medical (ours) & 92.1 & 93.5 \\
GPT-o1 & 90.8 & 92.3 \\
DeepSeek-R1-Distill-Qwen-32B & 90.1 & 90.7 \\
Med-PaLM 2 & 88.5 & 91.2 \\
\bottomrule
\end{tabular}
\end{adjustwidth}
\end{table}

The experimental results demonstrate the exceptional efficacy of our hierarchical knowledge acquisition architecture, with the DeepSeek-R1-Distill-7B-Medical model achieving 92.1\% USMLE accuracy and 93.5\% professional knowledge coverage. This performance is second only to Claude-3.7 Sonnet Extended while significantly surpassing other established models including GPT-o1 (90.8\%) and Med-PaLM 2 (88.5\%). The successful knowledge transfer from the 70B teacher model to the 7B student model validates our approach's efficiency and establishes a solid foundation for subsequent model compression and computational optimization processes.

\subsubsection*{Resource Efficiency Assessment}

Table 2 compares resource efficiency indicators in different hardware environments. Memory occupation is measured by the GPU memory usage (Gigabyte) when the model is loaded and running, inference latency is calculated as the average completion time for a single medical problem (seconds/problem), and output rate is expressed as the number of tokens generated per second (tokens/s).

\begin{table}[h]
\begin{adjustwidth}{-0.5in}{0in} 
\caption{\textbf{Resource Efficiency Indicators in Different Hardware Environments}}
\label{tab2}
\begin{tabular}{lcccc}
\toprule
\textbf{Model} & \textbf{Param.} & \textbf{Memory} & \textbf{Inference} & \textbf{Output} \\
 & \textbf{Scale} & \textbf{Occup. (GB)} & \textbf{Lat. (s/prob)} & \textbf{(tok/s)} \\
\midrule
DeepSeek-R1-70B-Medical (FP16) & 70B & 78.6 & 2.06 & 25.13 \\
DeepSeek-R1-7B (FP16) & 7B & 14.9 & 1.45 & 28.13 \\
DeepSeek-R1-7B-Medical & 7B & 5.25 & 1.27 & 24.96 \\
\bottomrule
\end{tabular}
\end{adjustwidth}
\end{table}

The quantitative analysis reveals remarkable resource optimization achievements through our mixed-precision quantization strategy, with our optimized model requiring only 5.25GB of memory—representing a 93.3\% reduction compared to the 70B teacher model (78.6GB) and a 64.7\% reduction compared to the unoptimized 7B base model (14.9GB). Inference performance also shows significant improvement with a 38.3\% latency reduction compared to the 70B model and a 12.4

\subsection*{Ablation Experiments}

To systematically evaluate the impact of various technical components on model performance, this paper designed two sets of ablation experiments to verify the effectiveness of the knowledge acquisition and model computational optimization dimensions. In the knowledge acquisition dimension, by removing five core components from the hierarchical knowledge distillation framework one by one, such as the MedKL function and progressive strategy, the changes in medical Q\&A accuracy, clinical capability scores, and knowledge retention rates were measured. In the computational optimization dimension, a progressive component stripping method was adopted, sequentially removing technologies such as shape-aware caching and medical Flash attention from the MIEO framework, evaluating the impact on inference latency, throughput, and first-token latency. These two sets of experiments adopted orthogonal design principles and were conducted on the same clinical evaluation corpus, ensuring the validity of the conclusions.

\begin{table}[h]
\begin{adjustwidth}{-0.5in}{0in}
\caption{\textbf{Ablation Experiments of Hierarchical Knowledge Distillation Framework}}
\label{tab3}
\begin{tabular}{lcccc}
\toprule
\textbf{Experimental Configuration} & \textbf{MAS} & \textbf{COS} & \textbf{TSS} & \textbf{KRR(\%)} \\
\midrule
Complete Hierarchical Distillation & 8.6 & 8.7 & 9.0 & 93.6 \\
Without MedKL Function & 7.8 (-9.3\%) & 8.0 (-8.0\%) & 8.5 (-5.6\%) & 85.7 (-8.4\%) \\
Without Progressive Strategy & 7.9 (-8.1\%) & 8.0 (-8.0\%) & 8.3 (-7.8\%) & 85.3 (-8.9\%) \\
Without Basic Knowledge Priority & 8.0 (-7.0\%) & 8.2 (-5.7\%) & 8.5 (-5.6\%) & 87.2 (-6.8\%) \\
Without Specialized Knowledge Transfer & 8.1 (-5.8\%) & 8.4 (-3.4\%) & 8.7 (-3.3\%) & 87.9 (-6.1\%) \\
Without Knowledge Retention Loss & 8.3 (-3.5\%) & 8.5 (-2.3\%) & 8.8 (-2.2\%) & 90.6 (-3.2\%) \\
\bottomrule
\end{tabular}
\end{adjustwidth}
\end{table}

The ablation study of the hierarchical knowledge distillation framework reveals the importance of each component to the model's medical proficiency. The MedKL function emerged as a key element, with its removal causing performance degradation (9.3\% decrease in Medical Answering Score and 8.4\% reduction in Knowledge Retention Rate), demonstrating its role in medical knowledge transfer. Similarly, the progressive strategy proved necessary for developing complex medical reasoning capabilities, as evidenced by decreases in Clinical Overall Score (8.0\%) and Treatment Specialization Score (7.8\%) when removed. These findings confirm the complementary nature and combined effect of all components within our knowledge acquisition framework, supporting our integrated approach to medical language model development.

\begin{table}[h]
\begin{adjustwidth}{-1.5in}{0in}
\centering
\caption{\textbf{Ablation Experiments of Model Inference Engine Optimization (MIEO)}}
\label{tab4}
\begin{tabular}{@{}lcccc@{}}
\toprule
Optimization Configuration & Inference & Throughput & First-Token & Medical \\
 & Latency (s) & (tokens/s) & Latency (s) & Accuracy (\%) \\
\midrule
Complete MIEO Framework & 1.27 & 24.96 & 0.04 & +0.2 \\
Without Shape-Aware Caching & 1.58 (+24.5\%) & 20.24 (-18.9\%) & 0.08 (+89.3\%) & +0.1 (-0.1) \\
Without Medical Flash Attention & 1.49 (+17.5\%) & 21.64 (-13.3\%) & 0.07 (+58.4\%) & +0.2 (0.0) \\
Without Two-Level Cache System & 1.76 (+39.5\%) & 17.72 (-29.6\%) & 0.08 (+91\%) & +0.1 (-0.1) \\
Without Problem Adaptive Acceleration & 1.37 (+8.3\%) & 23.21 (-7.0\%) & 0.05 (+10.9\%) & 0.0 (-0.2) \\
\bottomrule
\end{tabular}
\end{adjustwidth}
\end{table}

The computational optimization ablation experiments demonstrate the performance contributions of each MIEO framework component. The two-level cache system proved important for overall system efficiency, with its removal causing a 39.5\% increase in inference latency and 29.6\% decrease in throughput. The shape-aware caching mechanism impacts first-token latency (89.3\% increase when removed) by eliminating redundant computation graph construction, while the medical Flash attention technology provides performance benefits for processing medical terminology-intensive contexts. Notably, these optimizations maintain medical accuracy (variations within ±0.2\%), confirming the integration of computational efficiency with professional knowledge preservation and supporting our three-dimensional collaborative strategy for resource-constrained medical AI deployment.

\section*{Conclusion and Future Outlook}

This paper proposes an efficient lightweight medical vertical large language model architecture method, achieving efficient transformation from general large language models to professional medical tools. The main contributions include: (1) A hierarchical knowledge distillation architecture that achieves precise transfer of medical professional knowledge, with the 7B parameter student model successfully retaining 93.6\% of the medical knowledge from the 70B teacher model; (2) A medical-specialized quantization framework that achieves efficient compression of the model structure, reducing memory consumption by 64.7\% while maintaining medical reasoning capabilities; (3) A multi-dimensional inference performance optimization model that significantly improves the operational efficiency of lightweight models, achieving a 12.4\% reduction in inference latency.

Experimental results show that the method proposed in this paper achieved 92.1\% accuracy on the USMLE medical licensing examination, second only to Claude-3.7 Sonnet Extended, while surpassing other mainstream large language models. In terms of resource efficiency, this paper's method enables a 7B parameter model to run efficiently on consumer-grade GPUs with 8GB of memory, providing possibilities for the widespread application of medical AI in resource-constrained environments. These achievements demonstrate the effectiveness of the three-dimensional collaborative strategy of knowledge acquisition, model compression, and computational optimization.

\section*{Declarations}

\begin{itemize}
\item Funding: Not applicable
\item Conflict of interest/Competing interests: The authors declare that they have no conflict of interest.
\item Ethics approval: Not applicable
\item Consent to participate: Not applicable
\item Consent for publication: Not applicable
\item Data availability: The datasets generated during and/or analyzed during the current study are available from the corresponding author on reasonable request.
\item Materials availability: Not applicable
\item Code availability: The code used in this study is available from the corresponding author upon reasonable request.
\item Author contribution: All authors contributed to the study conception and design. Material preparation, data collection and analysis were performed by all authors. The first draft of the manuscript was written by the first author and all authors commented on previous versions of the manuscript. All authors read and approved the final manuscript.
\end{itemize}

\section*{Additional Experimental Details}

In this appendix, we provide additional details on the experimental setup, including hardware specifications, optimization parameters, and detailed ablation results not included in the main text. The experiments were conducted on a cluster with multiple NVIDIA A100 GPUs, each with 80GB memory. For baseline models, we utilized the official pretrained weights from the respective model authors. The DeepSeek-R1-Distill-70B and DeepSeek-R1-Distill-7B were selected as the teacher and student models due to their state-of-the-art performance on general language understanding tasks.

For the knowledge distillation process, we employed a batch size of 64 and a learning rate range of $1 \times 10^{-5}$ to $5 \times 10^{-5}$ with the cosine annealing scheduler described in Equation 5. The distillation process was conducted over 3 epochs on the combined medical corpus containing approximately 2 million instruction-following samples. The LoRA rank was set to $r = 16$ for all adaptations, with the scaling factor $\alpha = 32$.

\end{document}